\documentclass[preprint,12pt,authoryear]{elsarticle}

\usepackage{amssymb}
\usepackage{amsthm}
\usepackage{hyperref}
\usepackage{xcolor}
\usepackage{amsmath}
\usepackage{mathtools}
\usepackage{amsfonts}
\usepackage{bbm}
\usepackage{bm}
\usepackage{multirow}
\usepackage{subfigure}
\usepackage{booktabs}

\begin{document}

\begin{frontmatter}



\title{Why does Knowledge Distillation Work? Rethink its Attention and Fidelity Mechanism}


\author[inst1]{Chenqi Guo\corref{cor1}}
\cortext[cor1]{Corresponding author.}
\ead{chenqiguo72@ncepu.edu.cn}

\author[inst1]{Shiwei Zhong}
\ead{zsw@ncepu.edu.cn}

\author[inst1]{Xiaofeng Liu}
\ead{liu_xf@ncepu.edu.cn}

\author[inst2]{Qianli Feng}
\ead{fengq@amazon.com}

\author[inst1]{Yinglong Ma}
\ead{yinglongma@ncepu.edu.cn}

\affiliation[inst1]{organization={Control and Computer Engineering, North China Electric Power University},
            addressline={No. 2 Beinong Road}, 
            city={Beijing},
            postcode={102206},
            country={PR China}}

\affiliation[inst2]{organization={Amazon},
            addressline={300 Boren Ave N}, 
            city={Seattle},
            postcode={98109}, 
            state={WA},
            country={USA}}

\begin{abstract}
Does Knowledge Distillation (KD) really work? Conventional wisdom viewed it as a knowledge transfer procedure where a perfect mimicry of the student to its teacher is desired. However, paradoxical studies indicate that closely replicating the teacher's behavior does not consistently improve student generalization, posing questions on its possible causes. 
Confronted with this gap, we hypothesize that diverse attentions in teachers contribute to better student generalization at the expense of reduced fidelity in ensemble KD setups.
By increasing data augmentation strengths, our key findings reveal a decrease in the Intersection over Union (IoU) of attentions between teacher models, leading to reduced student overfitting and decreased fidelity. We propose this low-fidelity phenomenon as an underlying characteristic rather than a pathology when training KD. This suggests that stronger data augmentation fosters a broader perspective provided by the divergent teacher ensemble and lower student-teacher mutual information, benefiting generalization performance.
These insights clarify the mechanism on low-fidelity phenomenon in KD. Thus, we offer new perspectives on optimizing student model performance, by emphasizing increased diversity in teacher attentions and reduced mimicry behavior between teachers and student. Codes are available at \url{https://github.com/zisci2/RethinkKD}
\end{abstract}

\end{frontmatter}


\section{Introduction}\label{sec:introduction}

Knowledge Distillation (KD) \citep{hinton2015distilling} is renowned for its effectiveness in deep model compression and enhancement, emerging as a critical technique for knowledge transfer. Previously, this process has been understood and evaluated through model fidelity \citep{NEURIPS2021_KD}, measured by the student model replication degree to its teachers. High fidelity, assessed by metrics like low averaged predictive Kullback-Leibler (KL) divergence and high top-1 agreement \citep{NEURIPS2021_KD}, have conventionally been used to assess the success of KD.

While fidelity has traditionally guided enhancements in model architectures, optimization, and training frameworks, repeated high-fidelity results corresponding to strong student performance seem to indicate that a high degree of mimicry between the student and teachers is desirable \citep{KDbyMimic, li2021knowledge, Lao_2023_ICCV}. 
Yet this notion was initially challenged in \citep{NEURIPS2021_KD}, which empirically shows that good student accuracy does not imply good distillation fidelity in self and ensemble distillation. 
However, though \citep{NEURIPS2021_KD} underscores their empirical findings on the low-fidelity phenomenon, they still believe that closely matching the teacher is beneficial for KD in terms of knowledge transfer. Further, they identify optimization difficulties as one key reason of student's poor emulation behavior to its teachers. 
Thus, this paradox highlights a need for further exploration on model fidelity and its mechanism in KD. 

Among factors in KD analysis, the attention map mechanism serves as a pivotal role in understanding the student-teacher interplay.
It is known that in ensemble learning, diverse models improve the overall performance, and one can check their diversities through looking into the attention maps. Nonetheless, whether we can take it granted to transferring this conclusion into the case of KD has not been systematically studied yet.
For example, \citep{TSANTEKIDIS2021193} empirically shows that diversifying teachers' learnt policies by training them in different subsets of learning environment, can enhance the distilled student performance in KD. Yet, a theoretical foundation is lack for doing so. 
And it would be intriguing to check the student-teacher fidelity under such circumstance, to see if diversifying teacher models in an ensemble consistently corresponds with low-fidelity as well. If so, one can devote model attention map diversities to explain the existing fidelity paradox. 
Thus in this paper, we utilize the Intersection over Union (IoU) \citep{Rezatofighi_2019_CVPR} of attention maps \citep{zhou2016cvpr} between different teacher models in ensemble KD to help elucidate the existing fidelity paradox. 

Following the investigation paradigm in \citep{NEURIPS2021_KD}, where the model fidelity variations were observed with different data augmentations, we adapt this paradigm to our case with a more cautious control over the degree of randomness in augmentation during ensemble KD training. By varying data augmentation strengths, measured by \textit{Affinity} \citep{Affinity2021}, 
we varied the model diversities trained on them.
Impacts not only on traditional metrics like student-teacher fidelity, but also on less-explored aspects of attention maps diversity between different teachers, and mutual information between student and teachers are witnessed.
Our empirical observations appear to challenge the traditional wisdom on the student-teacher relationship in distillation during training procedure and thus provide further insights on explaining the fidelity paradox. 

Specifically, in support and further complement to \citep{NEURIPS2021_KD}, we highlight attention map diversification existed within teacher ensembles as a deeper reason why a student with good generalization performance may be unable to match the teacher during KD training: Stronger data augmentation increases attention divergence in the teacher ensemble, enabling teachers to offer a broader perspective to the student. Consequently, the student surpassing the knowledge of single teacher becomes more independent as measured by lower student-teacher mutual information. And the low-fidelity observed is a demonstration of this phenomenon.

Furthermore, though \citep{NEURIPS2021_KD} has demonstrated the low-fidelity observation, they still proposed the difficulties in optimization as the primary reason for it. And recent works including \citep{Sun2024Logit} remain optimizing in the direction of facilitating the student-teacher emulation procedure. Yet our empirical and theoretically analysis demonstrate that, optimization with logits matching does improve the student generalization ability but is still at the cost of fidelity reduction.


\begin{figure}[ht!]
\vskip 0.2in
\begin{center}
\centerline{\includegraphics[width=\textwidth]{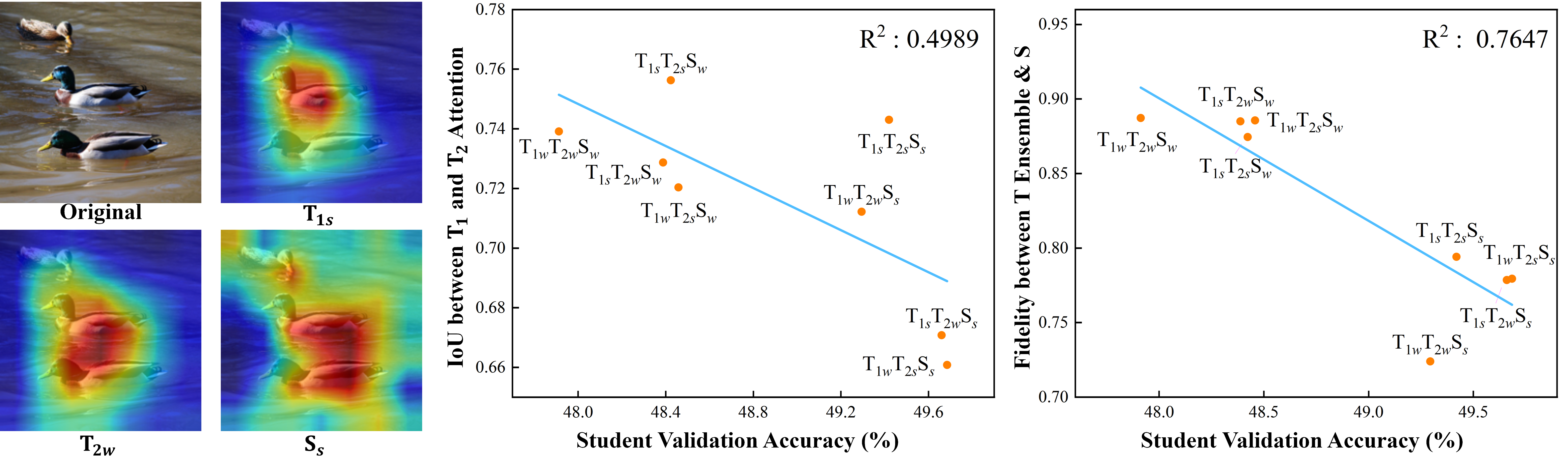}}
\caption{\textit{Left}: Attention map visualizations for teacher ensembles and student model in Knowledge Distillation (KD) on ImageNet dataset. Stronger data augmentation ($\text{T}_{1w}\text{T}_{2s}\text{S}_s$ and $\text{T}_{1s}\text{T}_{2w}\text{S}_s$ in this case) as measured by \textit{Affinity} improves teachers' attentional divergence, thus providing the student a more comprehensive perspective on the overall characteristics of the target images, leading to a better generalization ability. \textit{Middle and Right}: Scatter plots of Intersection over Union (IoU) in Attention maps, and Fidelity between teacher ensembles and student during KD training. The decreasing tendency in fidelity challenges the conventional wisdom that higher fidelity consistently correlate with better student performance. Later we will demonstrate that the low-fidelity observation is caused by attention map diversification existed within teacher ensembles, and even optimization towards logits-matching can hardly mitigate this low-fidelity effect.}
\label{fig:teaser}
\end{center}
\vskip -0.2in
\end{figure}

Our primary goal is to explain the fidelity paradox and understand the student learning and knowledge transfer dynamics in ensemble KD, by observing the implications of data augmentation on the student-teacher relationship.
By doing so, we seek to provide insights that challenge the traditional or extend preliminary wisdom in KD fidelity by leveraging the attention mechanism in ensemble learning. 
As shown in Figure \ref{fig:teaser}, we summarize our contributions as follows:

\begin{enumerate}
    \item[(1)] We demonstrate the correlation between teachers' attention map diversity and student model accuracy in ensemble KD training. Stronger data augmentation improves attentional divergence among teacher models, offering the student a more comprehensive perspective. 

    \item[(2)] We affirm the viewpoint from \citep{NEURIPS2021_KD} that higher fidelity between teachers and student does not consistently improve student performance. What is more, through analyzing attention maps between teachers in ensemble KD, we highlight this low-fidelity phenomenon as an underlying characteristic rather than a pathology: Student's generalization is enhanced with more diverse teacher models, which causes the reduction in student-teacher fidelity. 

    \item[(3)] We further investigate if optimization towards facilitating the student-teacher logits matching procedure can enhance the KD fidelity. Our empirical and theoretically analysis demonstrate that such optimization improve the student generalization ability but still at the cost of fidelity reduction.
\end{enumerate}

The rest of the paper is structured as follows: Section \ref{sec:related} summarizes the related works, Section \ref{sec:PS} clarifies the problem and hypothesis focused in this work, and Section \ref{sec:EMH} introduces the evaluation metrics used to validate our argues. Section \ref{sec:ES} further gives the experimental settings, and the empirical results and theoretical analysis are provided in Section \ref{sec:RA}. Section \ref{sec:conclusion} finally summarizes the work of this paper.

\section{Related Works}\label{sec:related}

Our study contributes to a growing body of research that explores the interactions between data augmentation, model fidelity, attention mechanisms, and their impact on student performance in Knowledge Distillation (KD) with teacher ensembles.

In \citep{DMAE2023}, a KD framework utilizing Masked Autoencoder, one of the primary factors influencing student performance is the randomness introduced by masks in its teacher ensembles.
It comes naturally if incorporating randomness into the dataset, through a simple yet effective method like data augmentation, and carefully controlling its strength, will be as effective as integrating it into model architectures.

Theories on the impacts of data augmentation on KD remain diverse and varied. \citep{Li_2022_ACCV} offers theoretical insights, suggesting that leveraging diverse augmented samples to aid the teacher model's training can enhance its performance but will not extend the same benefit to the student. 
\citep{PMLRshen22a} emphasizes how data augmentation can alter the relative importance of features, making challenging features more likely to be captured during the learning process. This effect is analogous to the multi-view data setting in ensemble learning, suggesting that data augmentation is likely to be beneficial for ensemble KD.

On the application font, research proposing novel attention-based KD frameworks usually accompanied with intricate designs in model architectures or data augmentation strategies \citep{OZDEMIR20226199, lewy2023attentionmix}. For instance, studies like \citep{Tian_2022} aim to address the few shot learning in KD with a novel data augmentation strategy based on the attentional response of the teacher model. Although their concentration is different from ours, the study nevertheless shows the significance of attention mechanism in KD.

In align with the initial ``knowledge transfer" definition of KD, as an underlying assumption that a higher degree of emulation between the student and teachers benefits its training, previous studies are devoted to optimizing towards increased student-teacher fidelity or mutual information \citep{KDbyMimic, li2021knowledge, Lao_2023_ICCV}. 
Recent work \citep{Sun2024Logit} also optimizes in this direction, where a z-score logit standardization process is proposed to mitigate the logits matching difficulties caused by logit shift and variance match between teacher and student. 
Nevertheless, this idea faced initial challenge in \citep{NEURIPS2021_KD}, indicating that closely replicating the teacher's behavior does not consistently lead to significantly improved student generalization performance during testing, whether in self-distillation or ensemble distillation.

\citep{NEURIPS2021_KD} first investigates if the low-fidelity is an identifiability problem that can be solved by augmenting the dataset, and the answer is no: experimental results show subtle benefits of this increased distillation dataset.
They further explore if the low-fidelity is an optimization problem resulting in a failure of the student to match the teacher even on the original training dataset, and their answer is yes: A shared initialization does make the student slightly more similar to the teacher in activation space, but in function space the results are indistinguishable from randomly initialized students.

Though insightful, it prompts further questions and drives us to think: Is low-fidelity truly undesirable and problematic for KD, especially if it does not harm student performance? Thus, additional exploration into this student fidelity-performance relation is required to elucidate the above paradox. Adopting a similar investigative approach which observes model fidelity variations with different data augmentations, we tailor it to our case, exercising a more cautious control over the data augmentation strength and thus the randomness into the distillation dataset during KD training.

In our work, we applied various data augmentations on KD, aiming to provide a more comprehensive understanding of model fidelity and attention mechanisms. Our empirical results and theoretical analysis challenge conventional wisdom, supporting and extending \citep{NEURIPS2021_KD} by demonstrating that student-teacher fidelity or mutual information does decrease with improved student performance during KD training. And, this low-fidelity phenomenon can hardly be mitigated with optimization aimed at improving student generalization. 
We thus advocate for more cautious practices in future research when designing KD strategies.

\section{Problem and Hypothesis}\label{sec:PS}

We focus on Knowledge Distillation (KD) with teacher ensembles in supervised image classification. In this realm, the efficacy of the process has traditionally been evaluated through the model fidelity and student validation accuracy. However, this conventional approach may not fully capture the complexity and nuances inherent in the knowledge transfer process, especially in light of evolving practices like data augmentation and the growing importance of attention mechanisms in neural networks. This study is driven by a series of interconnected research questions that challenge and extend the traditional understanding of KD as follows.

\textbf{Impact of Varied Data Augmentation Strengths on Model Diversity in Attention Map Mechanisms.}
The application of diverse data augmentation strengths during the training of teacher and student models plays a crucial role in shaping KD \citep{NEURIPS2021_KD}. Consequently, it is natural to inquire whether, across augmentation strategies, stronger data augmentation results in an increase or decrease in model fidelity within teacher ensembles during training. And if so, how does this correlate with the student model's performance. Inspired by the theory in machine learning that diversity among models can enhance ensemble learning performance \citep{EnsemMeth_Zhou2012, Asif2019EnsembleKD}, our hypothesis is that varying augmentation strengths in different teachers
inject randomness into the data, thereby diversifying teacher models' attention \citep{zhou2016cvpr} mechanisms trained on them. This diversity promotes heterogeneity in learning features, enables the student to learn diverse solutions to the target problem, and thus enhances the KD process. As a result, the student surpasses the knowledge of a single teacher, 
leading to a better overall performance, and the observed low-fidelity serves as a demonstration of this phenomenon.

\textbf{Interplay Between Student Fidelity, Mutual Information and Generalization.}
\citep{NEURIPS2021_KD, MiKD2023} have observed that fidelity or mutual information between teacher and student models interact with varying data augmentation strengths, influencing the overall effectiveness of distilled knowledge. The critical questions then arise: 
Does lower or higher fidelity and mutual information benefit the KD training and student performance, and why does it happen?
We hypothesize that, varied augmentation strengths in different teachers in ensemble KD would provide a broader view for the student to learn. Thus, the student surpassing the knowledge of a specific teacher. 
Contrary to the traditional perspective, we expect a decreased mimicry behavior of the student to benefit the student generalization ability during training, as it learns more intricate patterns from the diverse set of teachers.

\textbf{Effect of Optimization towards Student-Teacher Logits Matching on Fidelity.}
Question also comes on why some works thought a high-fidelity is beneficial, while others thought a low-fidelity is inevitable during training. 
Our intuition is that the researches devoted to optimizing towards increased student-teacher fidelity or mutual information do achieve the ultimate goal of improving the overall student performance, but in fact fail at enhancing the mimicry behavior during training.
In this paper, we try to answer this question by delving into a logits matching KD case as in \citep{Sun2024Logit}. Specifically, we experiment with a z-score standardization method to mitigate the logits magnitudes and variance gap between teacher and student, which facilitates the student-teacher emulation procedure. Our hypothesis is that though such an optimization can relieve the logit shift and variance match problem, in reality its benefit lies in the student generalization rather than the fidelity improvement. 

These questions aim to dissect the underlying learning dynamics in KD, moving beyond traditional metrics and exploring how newer facets like data augmentation strength, attention map diversity, fidelity and mutual information interplay to influence the student’s learning and generalization abilities.
Here, the data augmentation strength is measured by Affinity \citep{Affinity2021}, the offset in data distribution between the original one and the one after data augmentation as captured by the student model, which we will talk more later. By addressing these questions, this study seeks to provide a more comprehensive understanding of KD.

\section{Evaluation Metrics}\label{sec:EMH}

This section introduces evaluation metrics aimed at quantifying the learning dynamics and thus explains the existing fidelity paradox of Knowledge Distillation (KD) with teacher ensemble training, particularly when subject to varied data augmentation strengths.

\subsection{IoU in Attention Maps}

To elucidate divergent attentional patterns within teacher ensembles, we examine their attention maps \citep{zhou2016cvpr} in ResNet \citep{he2016resnet} or Transformer \citep{vas2017transf} during the training and validation stage. Subsequently, the Intersection over Union (IoU) \citep{Rezatofighi_2019_CVPR} is computed between the attention maps of different teachers to measure their diversities. Take the 2-teacher ensemble KD as an example, for an image sample $\boldsymbol{S}$, to compute the IoU between the teacher models, two attention maps $A_{t1}, A_{t2} \subseteq \boldsymbol{S}$ are obtained associated with each teacher model, with the final metric value computed as in Equation \ref{eq:IoU}:

\begin{equation}
    \text{IoU}=\frac{|A_{t1} \cap  A_{t2}|}{|A_{t1} \cup  A_{t2}|}
    \label{eq:IoU}
\end{equation}

\subsection{Model Dependency in KD} 

We use \textit{fidelity} metrics, namely the averaged predictive Kullback-Leibler (KL) divergence and top-1 agreement \citep{NEURIPS2021_KD}, along with mutual information calculated between models' logits. This enables us to showcase the mimicry behavior and dependency 
between teachers and the student.

Given a classification task with input space $\mathcal{X}=\{\bm{x}_i\}^N_{i=1}$ and label space $\mathcal{Y}=\{y_c\}^C_{c=1}$. Let $f:\mathcal{X} \to \mathbb{R}^C$ be a classifier whose outputs define a categorical predictive distribution over $\mathcal{Y}$, $\hat{p}(y_c|\bm{x}_i)= \sigma_c(\bm{z}_i)$, where $\sigma_c(\cdot)$ is the softmax function and $\bm{z}_i:=f(\bm{x}_i)$ denotes the model \textit{logits} when $\bm{x}_i$ is feed into $f$. The formal definition of KL divergence, top-1 agreement (Top-1 A), and mutual information (MI) are formulated as follows:

\begin{equation}
    \text{KL}(P_t||P_s) = \sum_{c=1}^C \hat{p}_t(y_c|\bm{x}) \log{\frac{\hat{p}_t(y_c|\bm{x})}{\hat{p}_s(y_c|\bm{x})}}
    \label{KL}
\end{equation}

\begin{equation}
    \text{Top-1 A} =  \frac{1}{N} \sum_{i=1}^{N} \mathbbm{1}\{ \text{arg}\max_c\sigma_c(\bm{z}^t)= \text{arg}\max_c\sigma_c(\bm{z}^s)\}
    \label{top1agreement}
\end{equation}

\begin{equation}
    \text{MI}(\mathcal{Y}^t;\mathcal{Y}^s) = \sum_{\bm{y}^t \in \mathcal{Y}^t}\sum_{\bm{y}^s \in \mathcal{Y}^s} P(\bm{y}^t,\bm{y}^s) \log{\frac{P(\bm{y}^t,\bm{y}^s)}{P(\bm{y}^t)P(\bm{y}^s)}}
    \label{MI}
\end{equation}
where $P(\bm{y}^t,\bm{y}^s)$ is the joint probability distribution of the teacher and student. $P(\bm{y}^t)$ and $P(\bm{y}^s)$ represent the marginal probability distributions of the teacher and student.
For metrics calculated between teach ensemble and student, the logits or outputs of different teachers are first averaged and then computed with the student.
This paper uses Top-1 A for fidelity measurement in the main text, and results with KL divergence can be found in \ref{app:KL_fidelity}.

\subsection{Quantify Data Augmentation Strength within Ensemble KD}

In our experiments, we employ various data augmentation techniques on both teacher ensembles and the student model to modulate the level of randomness introduced into the dataset, as detailed in Section \ref{sec:ES}. To quantify the strength of these applied data augmentations and demonstrate their effects on KD, we leverage \textit{Affinity} measurements \citep{Affinity2021}, specifically adapted to our KD scenario:

\begin{equation}
	\text{Affinity} = \frac{\text{Acc}(D'_{val})}{\text{Acc}(D_{val})} 
        \label{eq:affinity}
\end{equation}
where Acc$(D'_{val})$ denotes the validation accuracy of the student model trained with augmented distillation dataset and tested on the augmented validation set. Acc$(D_{val})$ represents the accuracy of the same model tested on clean validation set.

This metric measures the offset in data distribution between the original one and the augmented one captured by the student model after KD training: Higher Affinity value corresponds to smaller offset between the data distributions.
In this paper, Affinity is used as a tool to quantify and thus help on controlling the degree of randomness injected into the distillation dataset. This provides us with a systematic approach to analyze how data augmentation interacts with KD generalization, fidelity, and attention mechanisms. We anticipate that when the data augmentation strength of the student model aligns with that of the teacher model, the Affinity will be higher. And, lower Affinity corresponds to stronger data augmentation, leading to higher student accuracy and better generalization performance. 

It is noteworthy that what we mean \textit{low Affinity} is a ``moderate low but cannot be as low as 0" notion: An Affinity of 0 presupposes a situation where the augmented data is so drastically different from the original that it no longer retains any of the original data's informative features, or the model has entirely failed to learn from the augmented data. 
Our claim that models with low Affinity can still exhibit good generalization performance is based on the understanding that these models, through diverse and challenging augmentationss, learn to abstract and generalize from complex patterns. This does not necessarily imply that an Affinity of 0, resulting from complete misalignment with the augmented data, is desirable or indicative of strong generalization. Instead, we suggest that moderate to low Affinity, within a range that indicates the model has been challenged but still retains learning efficacy, can foster robustness and generalization.

\section{Experimental Setup}\label{sec:ES}

In our ensemble Knowledge Distillation (KD), experiments are conducted with two or three teachers. Each teacher model is a ResNet50 classifier pretrained on ImageNet \citep{deng2009imagenet} and then fine-tuned on their respective target datasets. The student model is ResNet18 trained from scratch using vanilla KD \citep{hinton2015distilling}. Take the ensemble KD with two teachers as an example, the loss function is defined as:

\begin{equation}
    \mathcal{L}_\text{NLL}(\bm{z}^s, \bm{y}^s)=-\sum_{c=1}^C y_c \log\sigma_c(\bm{z}^s)
\end{equation}

\begin{equation}
    \mathcal{L}_\text{KD1,2}(\bm{z}^s, \bm{z}^{t1.2})=-\tau^2\sum_{c=1}^C\sigma_c(\frac{\bm{z}^{t1,2}}{\tau})\log\sigma_c(\frac{\bm{z}^s}{\tau})
\end{equation}

\begin{equation}
    \mathcal{L}=\mathcal{L}_\text{NLL}+\frac{1}{2}(\mathcal{L}_\text{KD1}+\mathcal{L}_\text{KD2})
\end{equation}
where $\mathcal{L}_\text{NLL}$ is the usual supervised cross-entropy between the student logits $\bm{z}^s$ and the one-hot labels $\bm{y}^s$. $\mathcal{L}_\text{KD1,2}$ is the added knowledge distillation term that encourages the student to match the teacher ensembles.

In this paper, we are focusing on ensemble KD with 2 teachers $T_1$ and $T_2$. Results with 3 teachers are discussed in \ref{app:numT_enKD}. 
We also provide experiments with Vision Transformers (ViTs) \citep{dosovitskiy2020vit} where the attention map can be obtained directly with the built-in attention module in \ref{app:ViTresult}.

Experiments are conducted on well-recognized long-tailed datasets ImageNet-LT \citep{openlongtailrecognition}, CIFAR100 \citep{CIFAR2009} with an imbalanced factor of 100, and their balanced counterparts. Hyperparameters remain consistent across experiments for each dataset. More detailed settings, including learning rates and temperatures, are provided in  \ref{app:detail_ES}.

In this paper, we distinguish between two types of data augmentation: (1) Weak data augmentation, encompassing conventional methods such as random resized crop, random horizontal flip, and color jitters. (2) Strong data augmentation, which includes RandAugment (RA) \citep{RandAugment2020} applied on the ImageNet-LT dataset and AutoAugment (AA) \citep{AutoAugment2019} applied on all other datasets. For denotation purposes, we use $T_s, S_s$ to represent teacher or student models trained with strong augmentation, while $T_w, S_w$ denote those trained with weak augmentation.

It is essential to highlight that technically, the strong data augmentation applied to both teacher ensemble and student model in KD does not necessarily result in the highest data augmentation strength, as measured by our Affinity metric (defined in Equation \ref{eq:affinity}). This will be shown and clarified further in Section \ref{subsec:attentionMap} Table \ref{tab:Acc_Affi}.
Therefore, in this study, we varied the data augmentation strengths in ensemble KD. Specifically, in the series of experiments conducted on each dataset, we utilized the entire permutation set of $\text{T}_w, \text{T}_s, \text{S}_w, \text{S}_s$ to construct trials (for example, $\text{T}_{1s}\text{T}_{2w}\text{S}_s$ is one trial denotation), and then computed their Affinity to quantify their data augmentation strength. In practice, for evaluation, we computed our metrics introduced in Section \ref{sec:EMH} on both the training set and validation set, considering each trial's corresponding data augmentation strength.

\section{Results and Analysis}\label{sec:RA}

Our comprehensive set of experiments has yielded several intriguing insights into the learning dynamics of Knowledge Distillation (KD) and explains the fidelity paradox through various data augmentation strengths. We particularly emphasize the roles of attention map diversity, model fidelity, and mutual information, as they interact with student performance in terms of top-1 accuracy and overfitting during both the training and validation procedures.

\subsection{Impact on Attention Map Diversity}\label{subsec:attentionMap}

\begin{figure*}[ht]
\vskip 0.2in
\begin{center}
\centerline{\includegraphics[width=\linewidth]{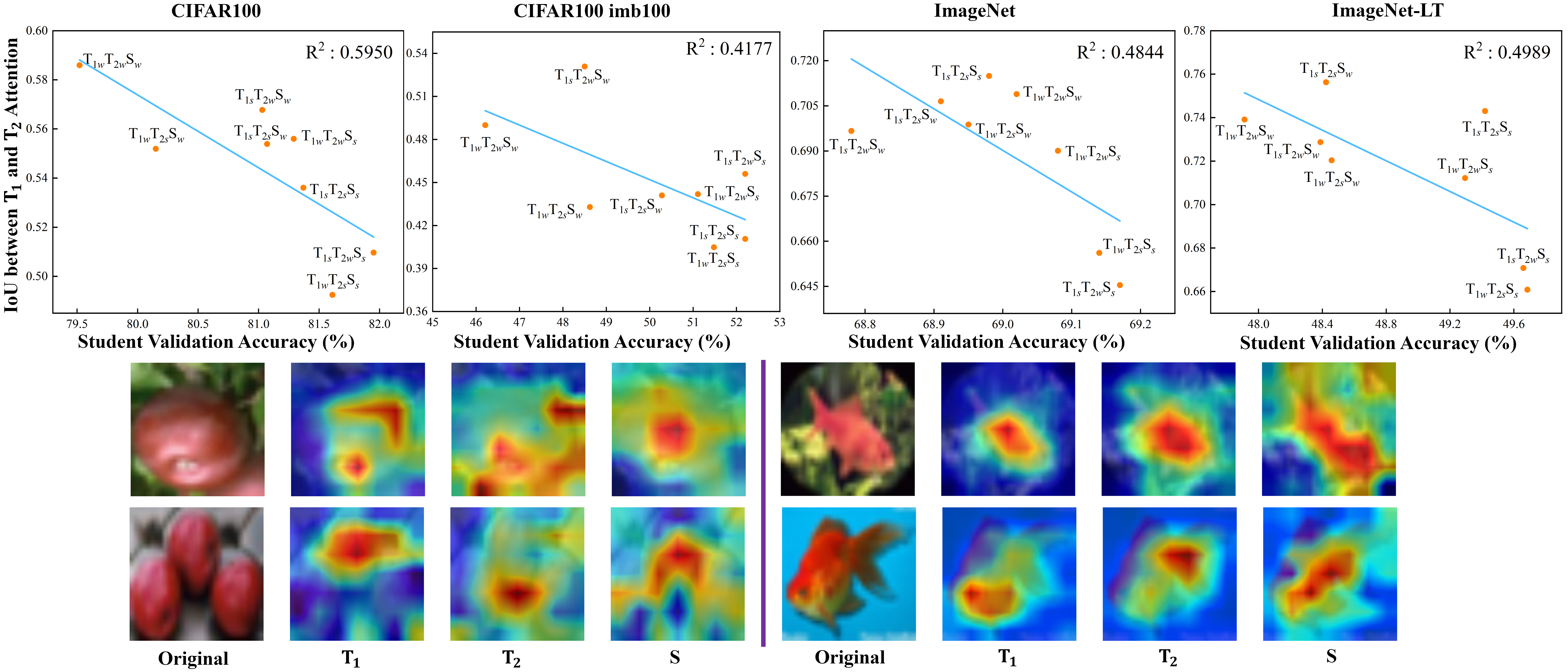}}
\caption{\textit{Top}: Scatter plots of IoU between $\text{T}_1$ and $\text{T}_2$ attention maps during KD training. \textit{Bottom}: Exampled attention maps of $\text{T}_1$, $\text{T}_2$ and S. This attention divergence among teacher ensembles, attributed to the randomness injected by data augmentation, gives the student distilled on them a more comprehensive perspective.} 
\label{fig:TT_IoU}
\end{center}
\vskip -0.2in
\end{figure*}

Figure \ref{fig:TT_IoU} \textit{Top} shows that during training, a consistent decrease is observed in the Intersection over Union (IoU) of attention maps between different teacher models with stronger data augmentation. This decrease is correlated with an increase in the student model’s accuracy. Trial denotations are also marked as data labels in these scatter plots, together with Table \ref{tab:Acc_Affi} demonstrating their data augmentation strengths. 


\begin{table*}[]
    \caption{Affinity, and Validation Accuracy (Val-Acc) of models with various data augmentation strengths.}
    \label{tab:Acc_Affi}
    \vskip 0.15in
    \begin{center}
    \begin{tiny}
    \begin{sc}
    \setlength{\tabcolsep}{1.05mm}{
    \begin{tabular*}{\linewidth}{lccccccccc} 
    \toprule
    \multirow{2}{*}{Dataset}                                                      & \multirow{2}{*}{Metric} & \multicolumn{8}{c}{Model}                                                             \\ \cmidrule(){3-10}
                                                                              &                         & T$_{1w}$T$_{2w}$S$_w$ & T$_{1w}$T$_{2w}$S$_s$ & T$_{1s}$T$_{2w}$S$_w$ & T$_{1s}$T$_{2w}$S$_s$ & T$_{1w}$T$_{2s}$S$_w$ & T$_{1w}$T$_{2s}$S$_s$ & T$_{1s}$T$_{2s}$S$_w$ & T$_{1s}$T$_{2s}$S$_s$ \\ \midrule
\multirow{2}{*}{Cifar100}                                                     & Affinity                & 0.9807   & \textbf{0.8611}   & 0.9805   & 0.9083   & 0.9858   & 0.9143   & 0.9729   & 0.9310   \\
                                                                              & Val-Acc            & 0.7952   & 0.8129   & 0.8103   & \textbf{0.8195}   & 0.8015   & 0.8161   & 0.8107   & 0.8137   \\ \midrule
\multirow{2}{*}{\begin{tabular}[c]{@{}c@{}}Cifar100\\ imb100\end{tabular}}    & Affinity                & 0.9763  & \textbf{0.8132}  & 0.9810  & 0.8637  & 0.9751  & 0.8635  & 0.9723  & 0.8955   \\
                                                                              & Val-Acc            & 0.4621   & 0.5111   & 0.4850    & \textbf{0.5220}   & 0.4862   & 0.5148   & 0.5028   & 0.5210   \\ \midrule
\multirow{2}{*}{ImageNet}                                                     & Affinity                & 0.9901   & \textbf{0.8767}   & 0.9930   & 0.8988   & 0.9845   & 0.9131   & 0.9871   & 0.9122   \\ 
                                                                              & Val-Acc            & 0.6902   & 0.6908   & 0.6878   & \textbf{0.6917}   & 0.6895   & 0.6914   & 0.6891   & 0.6898   \\\midrule
\multirow{2}{*}{\begin{tabular}[c]{@{}c@{}}ImageNet\\ long-tail\end{tabular}} & Affinity                & 0.9850   & \textbf{0.8311}   & 0.9755   & 0.8704   & 0.9782   & 0.8751   & 0.9903   & 0.8971   \\
                                                                              & Val-Acc            & 0.4791   & 0.4929   & 0.4839   & 0.4966   & 0.4846   & \textbf{0.4968}   & 0.4842   & 0.4942   \\
    
    \bottomrule 
    \end{tabular*}
    }
    \end{sc}
    \end{tiny}
    \end{center}
    \vskip -0.1in
\end{table*}

These Affinity values aid in understanding the data augmentation strengths and the decreasing tendencies in the scatter plots: Recall that Affinity measures the offset in data distribution between the original one and the one after data augmentation captured by the student, and lower Affinity corresponds to higher augmentation strength, leading to higher student accuracy. 
As evidence, for those trials with strong data augmentation and low Affinity, e.g., T$_{1s}$T$_{2w}$S$_s$ in CIFAR-100, T$_{1w}$T$_{2s}$S$_s$ in CIFAR-100 imb100, T$_{1s}$T$_{2w}$S$_s$ in ImageNet, and T$_{1s}$T$_{2w}$S$_s$ in ImageNet-LT, a relatively high validation accuracy is observed for each dataset.
It is important to emphasize that the application of strong data augmentation to both teacher ensemble and student model in KD does not lead to the highest level of data augmentation strength, as quantified by our Affinity metric defined in Equation \ref{eq:affinity}. 
That is, it is the diversity of teachers’ augmentation strength but not the strong data augmentation for a single teacher or student model matters: T$_{1s}$T$_{2w}$S$_s$ is stronger than T$_{1s}$T$_{2s}$S$_s$.
\ref{app:TTIoU_Affi} also offers scatter plots of IoU between $\text{T}_1$ and $\text{T}_2$ attention maps versus Affinity during KD training. 

Significantly, this observation suggests that as the ensemble of teachers focuses on increasingly diverse aspects of the input data, the student model benefits from a richer, more varied set of learned representations, leading to enhanced performance, as visualized in Figure \ref{fig:TT_IoU} \textit{Bottom}. 
This finding aligns with and extends ensemble learning theories in KD, where diversity among models enhances overall student performance even by simply manipulating the data augmentation strength. It introduces a new dimension to Knowledge Distillation theory, emphasizing the value of diverse learning stimuli.

\subsection{Revisiting the Role of Fidelity and Mutual Information}\label{subsec:fidMi}

\begin{figure*}[ht]
\vskip 0.2in
\begin{center}
\centerline{\includegraphics[width=\linewidth]{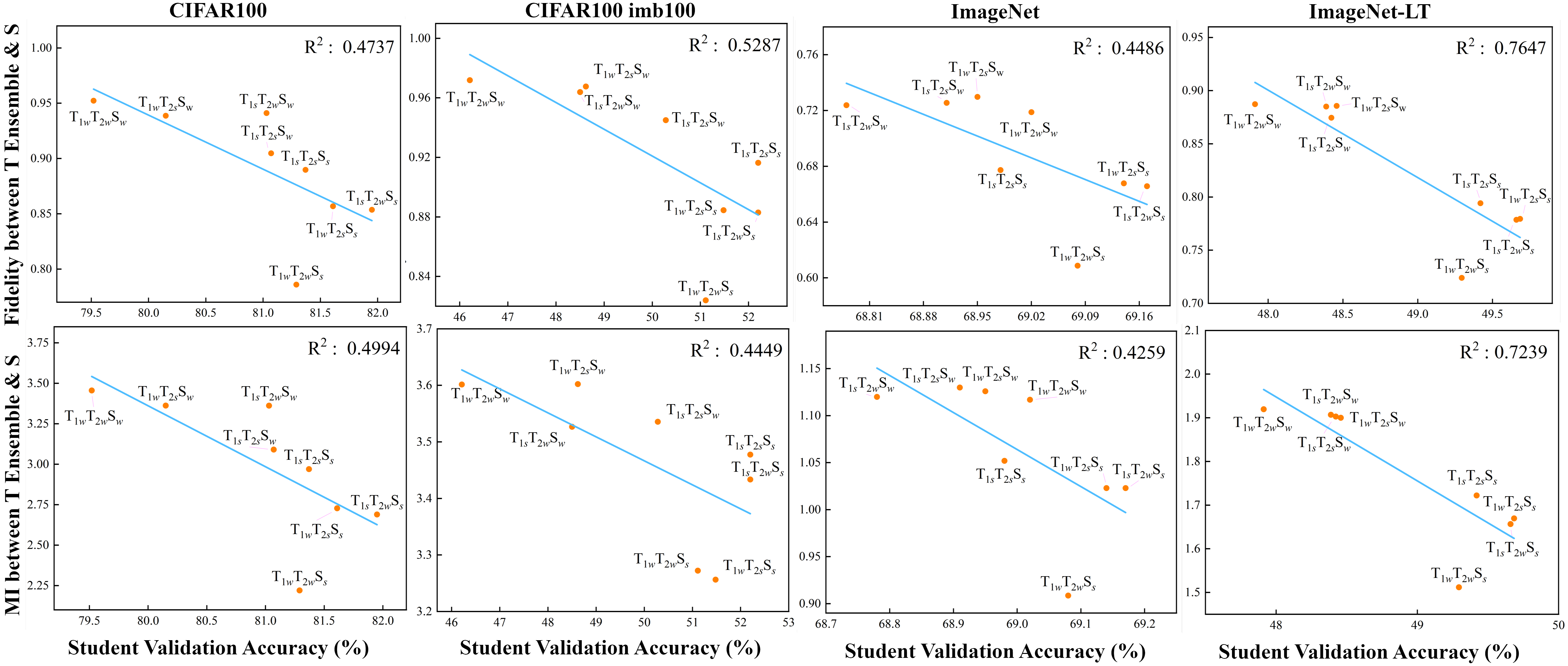}}
\caption{Scatter plots of \textit{Top}: Fidelity (measured by top-1 A) and \textit{Bottom}: Mutual Information (MI) between teacher ensembles and student during KD training. These decreasing tendencies along with the improved student validation accuracy are in contrast to the traditional viewpoint that higher fidelity consistently benefits student performance, indicating that some extent of student independency may be desired during KD training.}
\label{fig:TS_f_mi}
\end{center}
\vskip -0.2in
\end{figure*}

As in Figure \ref{fig:TS_f_mi}, during training, we observed a decrease in both fidelity and mutual information between teacher ensembles and the student model with stronger data augmentation. Intriguingly, this decrease was accompanied by improved validation accuracy in the student model. This indicates that a lower level of direct mimicry, 
in terms of output logits distribution, between teacher ensembles and the student is conducive to more effective learning in KD, possibly due to student learning from more divergent teachers' attentions. 


To further demonstrate the causality between teachers' attention divergence and low student-teacher fidelity, i.e., a more diverse attention maps within teacher ensemble causes a lower fidelity, an A/B test is conducted in the setup of ensemble KD with two teachers. 
Specifically, the control group is the vanilla KD (denoted as vKD) with different data augmentation strengths we used in all previous experiments, and the experimental group (denoted as hKD) is designed as follows: Each training image is first cropped into two parts, left and right, as input to teacher model $T_1$ and $T_2$ respectively. This allows us to proactively diversify the attention maps of each teacher model, rather than passively altering it in the case of varying data augmentation strengths.
Then in average, we can expect the experimental group to have far less attention IoU values than the control group, while keeping comparable generalization performance, because in the former each teacher's attention is constrained to one half of each image. The null hypothesis $H_0$ is that from control (vKD) to experimental (hKD) group, as the teachers' attention maps IoU decrease, an increase in student-teacher fidelity is observed. Denoting the total number of trials as $Num$, the corresponding $p$-value is calculated as:

\begin{equation}
    p\text{-value}=\frac{\#|\text{fidelity}(\text{hKD}) > \text{fidelity}(\text{vKD})|}{Num}
    \label{eq:pval}
\end{equation}

Experiments reveal a $p$-value less than 0.05, suggesting that we should reject this null hypothesis. Detailed experimental results are provided in \ref{app:attenLeftRightABtest}.
In summary, more divergent teacher attentions (i.e., lower IoU values) does cause the decrease in student-teacher fidelity.

This counterintuitive result aligns with and complements the paradoxical observation in \citep{NEURIPS2021_KD}. 
It implies that while the student model develops a certain level of independence from the teachers (evidenced by lower fidelity and mutual information), it still effectively captures and generalizes the core knowledge of the teachers. Combining with the observation on how varying data augmentation strengths influence the teachers' attention divergence in Section \ref{subsec:attentionMap}, we highlight attention diversification in teacher ensembles as a deeper reason why a student with good generalization may be unable to match the teacher during KD training: Stronger data augmentation increases attention divergence, enabling teachers to offer a broader perspective to the student. Consequently, the student surpasses the knowledge of a single teacher, becoming more independent, and the observed low-fidelity is a demonstration of this phenomenon rather than a pathology.

\subsection{Effects of Logits Matching Optimization on KD}\label{subsec:optim}

Although \citep{NEURIPS2021_KD} has shown the phenomenon of low-fidelity, they attributed the challenges in optimization as the key factor for the student's inability to match the teacher. Recent studies, such as \citep{Sun2024Logit}, continue to focus on optimizing the student-teacher logits matching process.
Yet in Section \ref{sec:PS} the 3rd hypothesis, we suggested that the optimization towards increasing student-teacher mimicry behavior in fact benefits generalization performance rather than the fidelity.

To illustrate, here we compared the aforementioned vanilla KD with a logits-matching optimization method in KD \citep{Sun2024Logit} under different data augmentation strengths, for dataset CIFAR100, CIFAR100-imb100, and ImageNet-LT. 
Specifically, we experiment with a z-score standardization method applied on logits before the softmax. This mitigates the logits magnitudes and variance gap between teacher and student, which facilitates the student-teacher emulation procedure. 

Theoretically, 
denote the logits of teacher model and student model as $\bm{z}^t$ and $\bm{z}^s$ respectively, and the softmax function as $\sigma(\cdot)$.
Then for a finally well-distilled student with predicted probability density perfectly matching the teacher, i.e., $\sigma(\bm{z}^s)=\sigma(\bm{z}^t)$, we have the following two properties proved in \citep{Sun2024Logit}:

\begin{equation}
    \text{Logit shift: } \bm{z}^s = \bm{z}^t+\Delta
    \label{eq:logitShift}
\end{equation}

\begin{equation}
    \text{Variance match: } \frac{\text{Var}(\bm{z}^s)}{\text{Var}(\bm{z}^t)} = \frac{\tau_s}{\tau_t}
    \label{eq:varMatch}
\end{equation}

Where $\Delta$ can be considered constant for each sample image, and $\tau_s, \tau_t$ are temperatures for the student and teacher respectively during training.
That is, even for the student with highest fidelity to its teacher such that $\sigma_c(\bm{z}^s)=\sigma_c(\bm{z}^t)$ for any class $c$ in the dataset, still we have $\bm{z}^s=\sqrt{\frac{\tau_s}{\tau_t}}\cdot\bm{z}^t+\Delta$ which means the student logits cannot match the teacher logtis. A z-score normalization applied on both the student and teacher logits during KD training can soothe this mismatch by making their logtis distribution equal mean and variance, and thus improve generalization performance.
However, from the fidelity definition in Equation \ref{top1agreement}, since the softmax function is monotonic, what we are looking for is the agreed index $c$ of maximum logits between the teacher and student $\text{arg}\max_c(\bm{z}^t)= \text{arg}\max_c(\bm{z}^s)$, which unfortunately cannot be directly affected by such optimization method.

\begin{figure*}[]
\vskip 0.2in
\begin{center}
\centerline{\includegraphics[width=\linewidth]{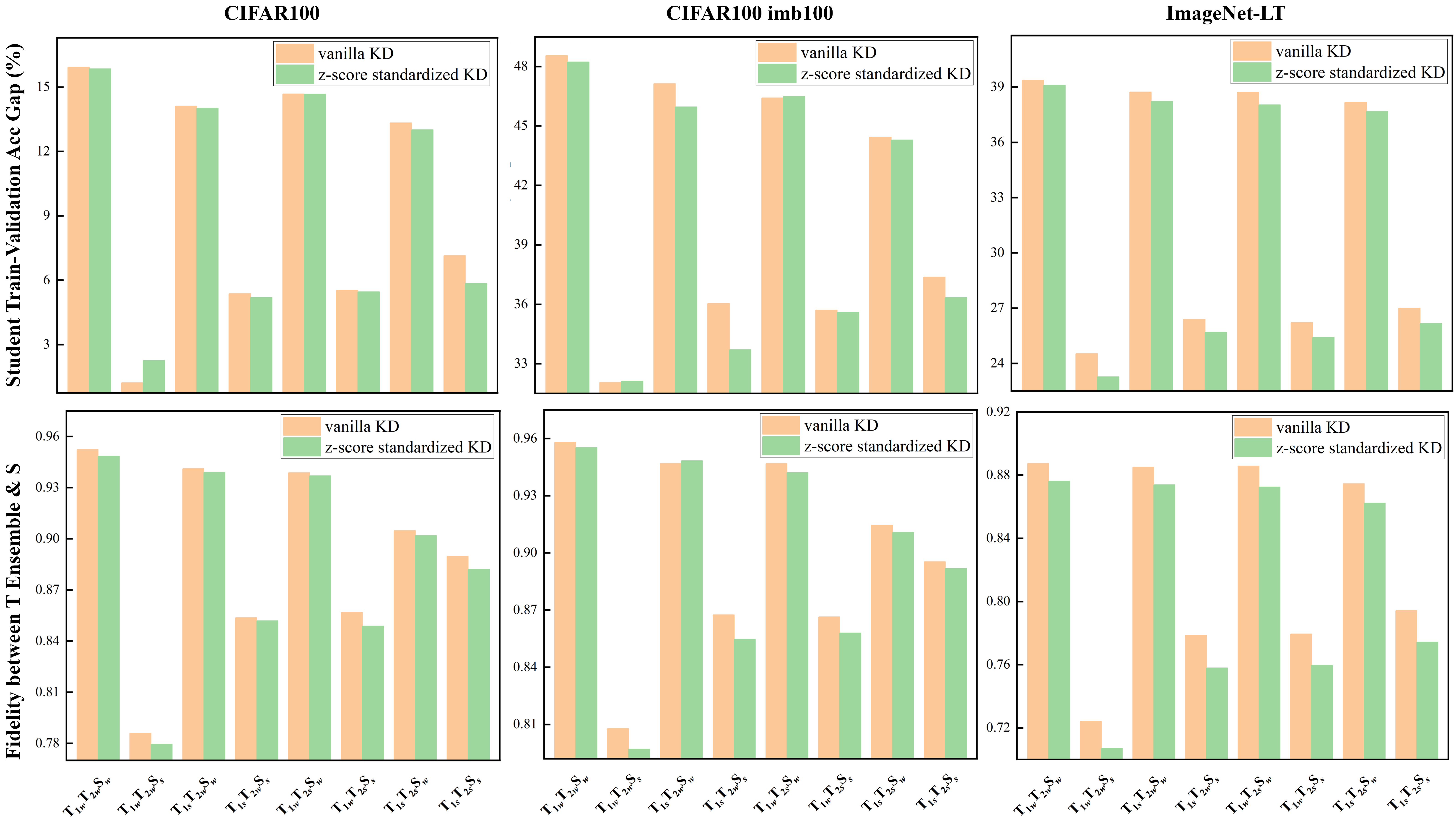}}
\caption{Bar plots comparing between vanilla KD and z-score standardization KD. \textit{Top}: Generalization performance in terms of train-validation accuracy gap. \textit{bottom}: Student-teacher fidelity. The z-score standardization, aimed at facilitating the student-teacher logits matching procedure, does improve student generalization performance (indicated by a lower accuracy gap) in most cases. However,  it also leads to a decrease in student-teacher fidelity during training, suggesting that the benefit lies more in student generalization than in fidelity improvement.}
\label{fig:zscoreNorm}
\end{center}
\vskip -0.2in
\end{figure*}

In conclusion, though an optimization towards student-teacher logits matching can relieve the logit shift and variance match problem, in reality its benefit lies in the student generalization rather than the fidelity improvement.
As shown in Figure \ref{fig:zscoreNorm}, the z-score standardization does improve the student train-validation accuracy gap in most cases, but a decrease in the student-teacher fidelity is still witnessed. 


\section{Conclusion}\label{sec:conclusion}

Our research, aiming to explain the fidelity paradox, intersects with and expands upon existing theories for ensemble Knowledge Distillation (KD) in several ways. \textbf{(1)} It introduces a novel perspective on the learning and knowledge transfer process by investigating the impact of attention map diversity on fidelity in KD with various data augmentation strength. \textbf{(2)} It reevaluates the teacher-student fidelity and mutual information challenge, providing insights into the ongoing debates about the relation between student's ability to mimic its teachers and its generalization performance in KD. \textbf{(3)} It highlights that for optimization towards facilitating student-teacher logits matching which relieves the logit shift and variance match problem, its benefit lies in the student generalization rather than the fidelity improvement.
These insights have the potential to catalyze further theoretical advancements in the pursuit of robust KD.

\appendix

\section{Detailed Experimental Settings}\label{app:detail_ES}

The experiments are run on a GPU machine with RTX 4090 GPU, AMD 5995WX CPU and 128 GB memory. 
In each trial, the teacher model of ResNet50 is trained for 30 epochs for ImageNet-LT dataset, and 60 epochs for all the others. 
The student model of ResNet18 is distilled for: 200 epochs for CIFAR-100; 175 epochs for CIFAR-100 imb100; 60 epochs for ImageNet; and 165 epochs for ImageNet-LT dataset, when their validation accuracy converges. 

Hyper-parameters, including temperatures of $\tau=10$, hard label weight of $\alpha=0.2$, initial learning rate of $0.1$, momentum of $0.9$, and batch size of $128$, remain the same throughout the entire procedure in each case, ensuring consistent and reliable results for evaluation.

For training with balanced ImageNet dataset, we use a cosine annealing learning rate scheduler, with $T_\text{max}=30, \text{eta}_\text{min}=0$ for teacher training, and $T_\text{max}=60, \text{eta}_\text{min}=0$ for student distillation. 
For other datasets, a lambda learning rate scheduler is used.
Specifically, during teacher training, with the following hyperparameters: $\text{step}_1=25, \text{step}_2=40,
\text{step}_3=60$ for CIFAR-100; $\text{step}_1=25, \text{step}_2=40, \text{step}_3=60$ for CIFAR-100 imb100; and $\text{step}_1=35, \text{step}_2=50$ for ImageNet-LT.
During student distillation, with the following hyperparameters: $\text{step}_1=190, \text{step}_2=195$ for CIFAR-100; $\text{step}_1=160, \text{step}_2=165, \text{step}_3=170$ for CIFAR-100 imb100; and $\text{step}_1=150, \text{step}_2=155, \text{step}_3=160$ for ImageNet-LT.

\section{Fidelity with KL divergence Measurement}\label{app:KL_fidelity}

\begin{figure*}[]
\vskip 0.2in
\begin{center}
\centerline{\includegraphics[width=\linewidth]{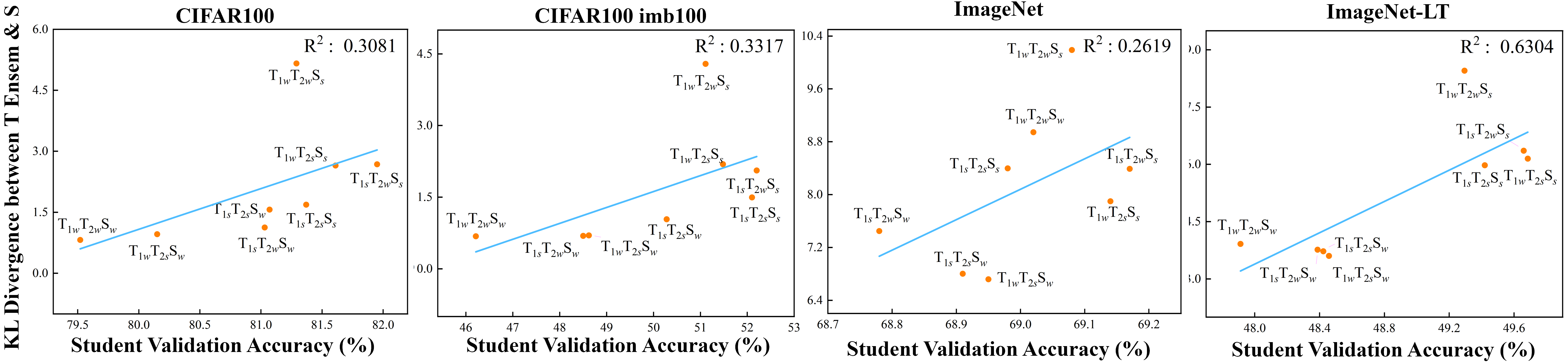}}
\caption{Scatter plots of fidelity (measured by KL divergence) between teacher ensembles and student during KD training. For KL divergence, a higher value implies lower fidelity. Thus, these increasing tendencies align with the decreasing ones with Top-1 A in the main text.}
\label{fig:APP_TS_KL}
\end{center}
\vskip -0.2in
\end{figure*}

In the main text Section \ref{subsec:fidMi}, Top-1 A is used for the fidelity metric. Here we also provide results with Kullback-Leibler (KL) divergence between teacher ensembles and student during KD training, as in Figure \ref{fig:APP_TS_KL}. Note that for KL divergence, a higher value implies lower fidelity.


\section{In-Depth Results for the A/B Test}\label{app:attenLeftRightABtest}

In the main text, to demonstrate the causality between teachers' attention divergence and low student-teacher fidelity, an A/B test is conducted for ensemble KD with two teachers. Experiments reveal a $p$-value less than 0.05, suggesting that more divergent teacher attentions (i.e., lower IoU values) does cause the decrease in student-teacher fidelity. 
In this section, we further provides the detailed experimental results of the A/B test, as shown in Table \ref{tab:vanilla_CIFAR100}, \ref{tab:vanilla_CIFAR100_IMB100} and \ref{tab:vanilla_ImageNetLT}. Here, vKD denotes the control group of vanilla KD experiments, and hKD denotes the control group of half-image inputs experiments. From these results, it can be seen that in average, hKD has far less attention IoU values than vKD, while keeping comparable generalization performance (indicated by a lower accuracy gap).


\begin{table*}[]
    \caption{Results for the A/B Test on CIFAR100 Dataset.}  
    \label{tab:vanilla_CIFAR100}
    \vskip 0.15in
    \begin{center}
    \begin{small}
    \begin{sc}
    \setlength{\tabcolsep}{3.5mm}{
    \begin{tabular*}{\linewidth}{lcccccc}
    \toprule
    \multirow{2}{*}{Model} & \multicolumn{2}{c}{Acc Gap} & \multicolumn{2}{c}{IoU} & \multicolumn{2}{c}{Fidelity} \\ \cmidrule(lr){2-3} \cmidrule(lr){4-5} \cmidrule(lr){6-7} 
                           & vKD          & hKD          & vKD       & hKD         & vKD           & hKD          \\ \midrule
    T$_{1w}$T$_{2w}$S$_w$  & 0.1593       & 0.1631       & 0.5860    & 0.3188      & 0.9523        & 0.7564       \\
    T$_{1w}$T$_{2w}$S$_s$  & 0.0122       & 0.0171       & 0.5560    & 0.3062      & 0.7859        & 0.5921       \\
    T$_{1s}$T$_{2w}$S$_w$  & 0.1411       & 0.1560       & 0.5678    & 0.3033      & 0.9411        & 0.7295       \\
    T$_{1s}$T$_{2w}$S$_s$  & 0.0537       & 0.0654       & 0.5097    & 0.2970      & 0.8536        & 0.6520       \\
    T$_{1w}$T$_{2s}$S$_w$  & 0.1468       & 0.1784       & 0.5519    & 0.2619      & 0.9387        & 0.7248       \\
    T$_{1w}$T$_{2s}$S$_s$  & 0.0553       & 0.0759       & 0.4925    & 0.2549      & 0.8568        & 0.6513       \\
    T$_{1s}$T$_{2s}$S$_w$  & 0.1333       & 0.1541       & 0.5539    & 0.2738      & 0.9048        & 0.6621       \\
    T$_{1s}$T$_{2s}$S$_s$  & 0.0714       & 0.0657       & 0.5361    & 0.2747      & 0.8897        & 0.6801       \\ \bottomrule 
    \end{tabular*}
    }
    \end{sc}
    \end{small}
    \end{center}
    \vskip -0.1in
\end{table*}

\begin{table*}[]
    \caption{Results for the A/B Test on CIFAR100 IMB100 Dataset.} 
    \label{tab:vanilla_CIFAR100_IMB100}
    \vskip 0.15in
    \begin{center}
    \begin{small}
    \begin{sc}
    \setlength{\tabcolsep}{3.5mm}{
    \begin{tabular*}{\linewidth}{lcccccc}
    \toprule
    \multirow{2}{*}{Model} & \multicolumn{2}{c}{Acc Gap} & \multicolumn{2}{c}{IoU} & \multicolumn{2}{c}{Fidelity} \\ \cmidrule(lr){2-3} \cmidrule(lr){4-5} \cmidrule(lr){6-7} 
                       & vKD          & hKD          & vKD       & hKD         & vKD           & hKD          \\ \midrule
    T$_{1w}$T$_{2w}$S$_w$  & 0.4854                  & 0.4836                  & 0.4900                  & 0.3195                  & 0.9580                  & 0.7114                  \\
    T$_{1w}$T$_{2w}$S$_s$  & 0.3206                  & 0.3742                  & 0.4419                  & 0.3094                  & 0.8078                  & 0.5406                  \\
    T$_{1s}$T$_{2w}$S$_w$  & 0.4712                  & 0.4995                  & 0.5309                  & 0.3041                  & 0.9467                  & 0.6892                  \\
    T$_{1s}$T$_{2w}$S$_s$  & 0.3604                  & 0.3994                  & 0.4560                  & 0.2992                  & 0.8675                  & 0.6040                  \\
    T$_{1w}$T$_{2s}$S$_w$  & 0.4641                  & 0.4860                  & 0.4329                  & 0.2643                  & 0.9467                  & 0.6892                  \\
    T$_{1w}$T$_{2s}$S$_s$  & 0.3570                  & 0.3827                  & 0.4084                  & 0.2558                  & 0.8664                  & 0.5997                  \\
    T$_{1s}$T$_{2s}$S$_w$  & 0.4444                  & 0.4790                  & 0.4410                  & 0.2717                  & 0.9145                  & 0.6192                  \\
    T$_{1s}$T$_{2s}$S$_s$  & 0.3738                  & 0.3225                  & 0.4107                  & 0.2721                  & 0.8953                  & 0.6242                  \\ \bottomrule 
    \end{tabular*}
    }
    \end{sc}
    \end{small}
    \end{center}
    \vskip -0.1in
\end{table*}

\begin{table*}[]
    \caption{Results for the A/B Test on ImageNet Long-tail Dataset.} 
    \label{tab:vanilla_ImageNetLT}
    \vskip 0.15in
    \begin{center}
    \begin{small}
    \begin{sc}
    \setlength{\tabcolsep}{3.5mm}{
    \begin{tabular*}{\linewidth}{lcccccc}
    \toprule
    \multirow{2}{*}{Model} & \multicolumn{2}{c}{Acc Gap} & \multicolumn{2}{c}{IoU} & \multicolumn{2}{c}{Fidelity} \\ \cmidrule(lr){2-3} \cmidrule(lr){4-5} \cmidrule(lr){6-7} 
                       & vKD          & hKD          & vKD       & hKD         & vKD           & hKD          \\ \midrule
    T$_{1w}$T$_{2w}$S$_w$  & 0.3937                  & 0.4104                  & 0.7391                  & 0.6245                  & 0.8873                  & 0.5657                  \\
    T$_{1w}$T$_{2w}$S$_s$  & 0.2453                  & 0.2426                  & 0.7122                        & 0.6311                  & 0.7240                  & 0.4542                  \\
    T$_{1s}$T$_{2w}$S$_w$  & 0.3873                  & 0.4152                  & 0.7287                        & 0.5948                  & 0.8850                  & 0.5554                  \\
    T$_{1s}$T$_{2w}$S$_s$  & 0.2639                  & 0.2713                  & 0.6708                  & 0.5607                  & 0.7786                  & 0.4901                  \\
    T$_{1w}$T$_{2s}$S$_w$  & 0.3871                  & 0.4161                  & 0.7204                  & 0.5798                  & 0.8856                  & 0.5559                  \\
    T$_{1w}$T$_{2s}$S$_s$  & 0.2622                  & 0.2680                  & 0.6608                  & 0.5537                  & 0.7795                  & 0.4916                  \\
    T$_{1s}$T$_{2s}$S$_w$  & 0.3816                  & 0.4133                  & 0.7563                  & 0.6244                  & 0.8745                  & 0.5308                  \\
    T$_{1s}$T$_{2s}$S$_s$  & 0.2700                  & 0.2663                  & 0.7431                  & 0.6490                  & 0.7941                  & 0.5138                  \\ \bottomrule 
    \end{tabular*}
    }
    \end{sc}
    \end{small}
    \end{center}
    \vskip -0.1in
\end{table*}




\section{IoU between $\text{T}_1$ and $\text{T}_2$ Attentions versus Affinity}\label{app:TTIoU_Affi}

\begin{figure*}[h]
\vskip 0.2in
\begin{center}
\centerline{\includegraphics[width=\linewidth]{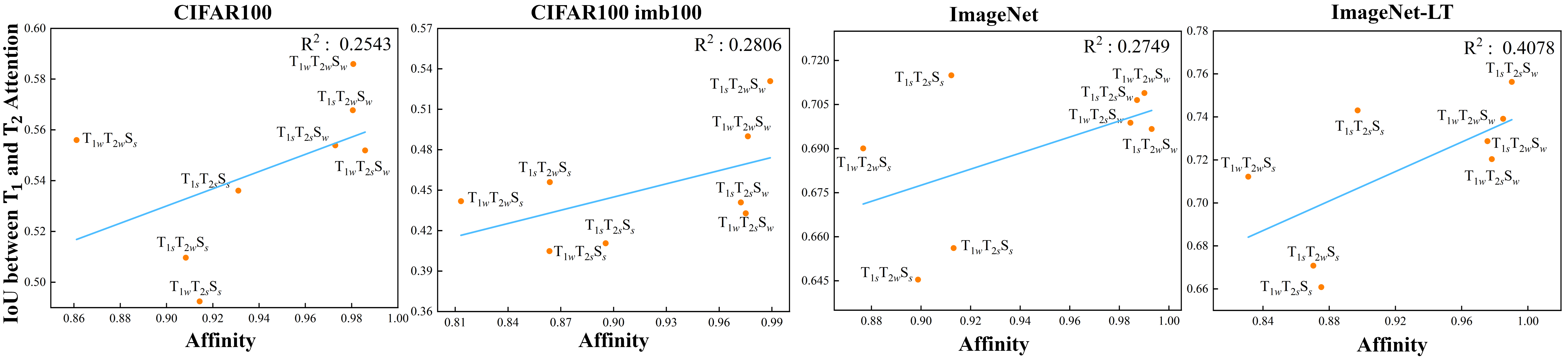}}
\caption{Scatter plots of IoU between $\text{T}_1$ and $\text{T}_2$ attention maps versus Affinity during KD training. These increasing tendencies demonstrate that stronger data augmentation (indicated by smaller Affinity) does correlate with more divergent teacher attentions (indicated by lower IoU).}
\label{fig:APP_TTIoU_Affi}
\end{center}
\vskip -0.2in
\end{figure*}

In the main text, we show that during training, a consistent decrease is observed in the Intersection over Union (IoU) of attention maps between different teacher models versus student validation accuracy, suggesting that more divergent teacher attentions correlate with higher accuracy. 
Here, we also provide the scatter plots of IoU between $\text{T}_1$ and $\text{T}_2$ attention maps versus Affinity during KD training, as in Figure \ref{fig:APP_TTIoU_Affi}. These increasing trends demonstrate that stronger data augmentation (indicated by smaller Affinity) does correlate with more divergent teacher attentions (indicated by lower IoU).


\section{Experiments with Vision Transformers}\label{app:ViTresult}

\begin{figure*}[h]
\vskip 0.2in
\begin{center}
\centerline{\includegraphics[width=\linewidth]{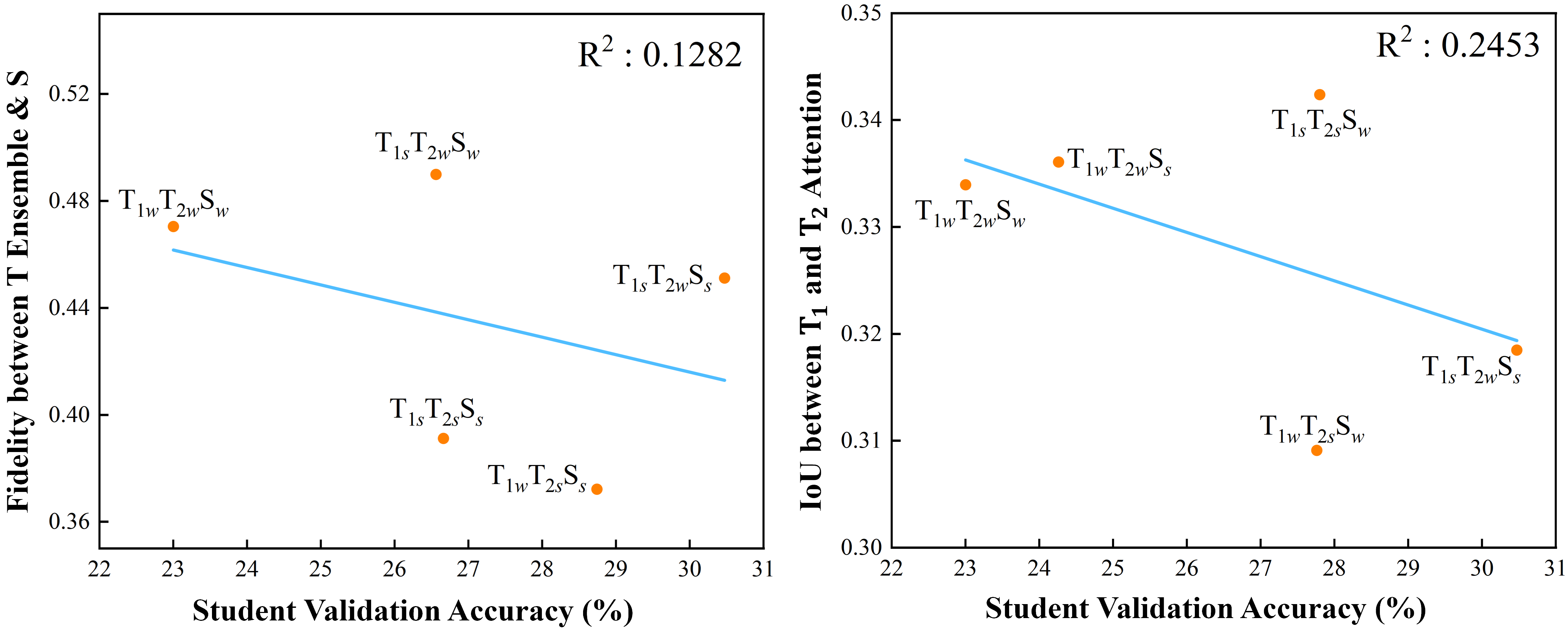}}
\caption{Scatter plots for experiments with Vision Transformer (ViT) on CIFAR100 imb100 dataset. \textit{Left}: Fidelity (measured by top-1 A) and \textit{Right}: IoU between $\text{T}_1$ and $\text{T}_2$ during KD training. These decreasing tendencies align with our conclusions drawn from ResNet experiments, suggesting the applicability of our analysis method to attention-based methods like ViT. The main distinction is in calculating IoU, where we can directly use ViT's built-in attention module to obtain the attention maps.}
\label{fig:APP_ViT}
\end{center}
\vskip -0.2in
\end{figure*}

In this section, we also provide experiments with Vision Transformers (ViTs) \citep{dosovitskiy2020vit} on CIFAR100 imb100 dataset where the attention map can be obtained directly with the built-in attention module.
As shown in Figure \ref{fig:APP_ViT}, our analysis method can be applied to attention-based methods such as ViT. The only difference is that when calculating IoU, we can directly use the built-in attention module of ViT to obtain the attention maps. In this experiment, two ViT-b32 teachers are distilled on one ViT-b16 student for CIFAR100 imb100 dataset. And the conclusions in our manuscript still holds for these two cases. That is, lower student-teacher fidelity and larger teachers’ attention diversity correlate with higher student validation accuracy.


\section{Results with More Teacher Numbers in Ensemble Knowledge Distillation}\label{app:numT_enKD}

\begin{figure*}[h]
\vskip 0.2in
\begin{center}
\centerline{\includegraphics[width=\linewidth]{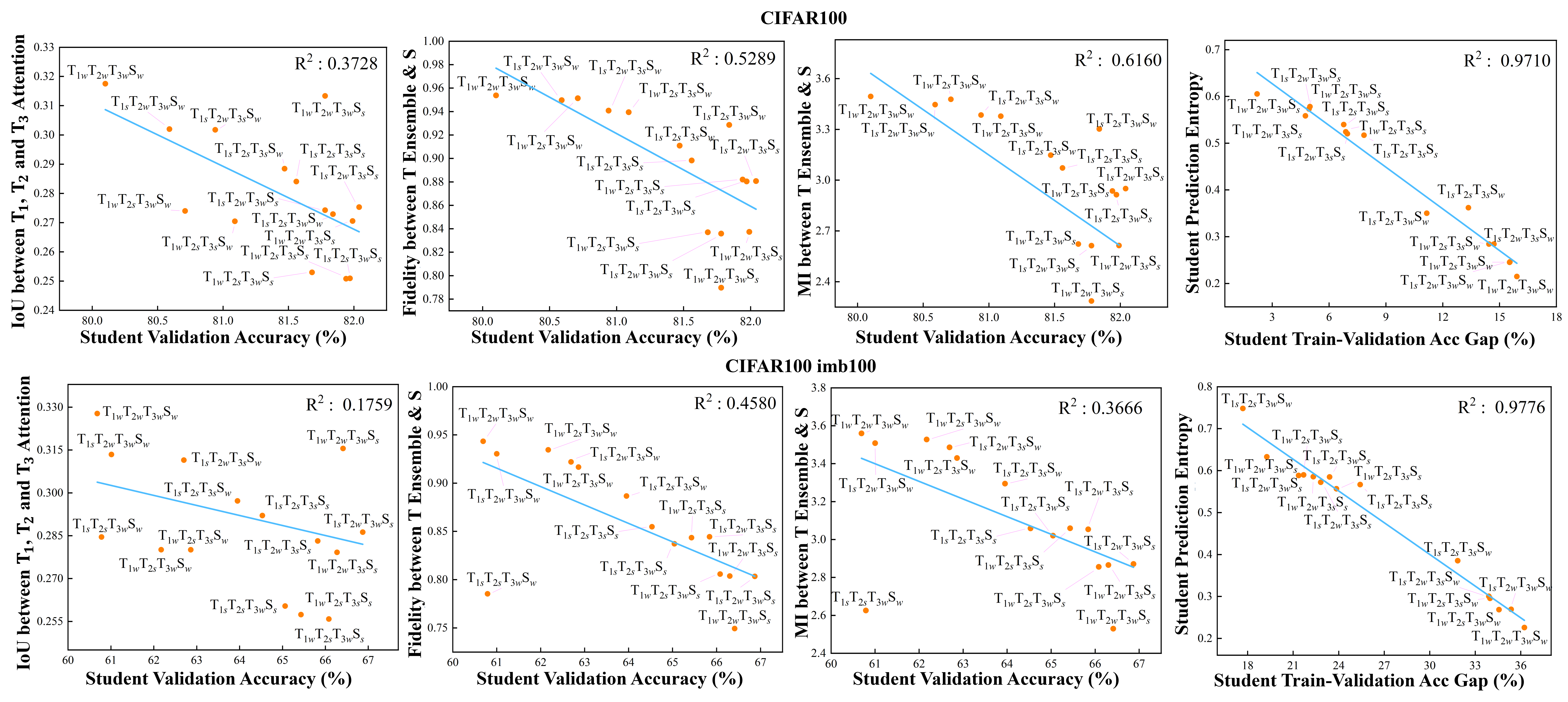}}
\caption{Scatter plots of teacher attention IoU, fidelity, mutual information, and student entropy in 3-teacher ensemble KD cases. These results, aligning with the tendencies observed in 2-teacher cases, further support our conclusions in the main text.}
\label{fig:APP_plts_3T}
\end{center}
\vskip -0.2in
\end{figure*}

In the main text, we focused on Knowledge Distillation (KD) with 2 teachers in the ensemble. Results with 3 teachers are discussed here. Figure \ref{fig:APP_plts_3T} provides scatter plots of teacher attention IoU, fidelity, mutual information, and student entropy in 3-teacher ensemble KD cases, for CIFAR100 and CIFAR100 imb100 datasets. These plots align with the tendencies observed in 2-teacher cases in the main text.


\section{Quantitative Evaluation}\label{app:quantEval}

Table \ref{tab:val_acc} compares our method with SOTA baselines: LFME \cite{LFME2020} and DMAE \cite{DMAE2023}, focusing on the top-1 validation accuracy. LFME is specifically designed for long-tailed datasets, so we only present its results on those. DMAE is initially designed for balanced datasets, so its performance on balanced ones is less satisfying.
For our method shown in this table: Ours(1T) is refferred to the KD with one ResNet50 teacher model distilled to one ResNet18 student model, with $\text{T}_{w}\text{S}_s$. Ours(2T) is refferred to the KD with two ResNet50 teacher models distilled to one ResNet18 student model, with $\text{T}_{1s}\text{T}_{2w}\text{S}_s$. Ours(3T) is refferred to the KD with three ResNet50 teacher models distilled to one ResNet18 student model, with $\text{T}_{1s}\text{T}_{2w}\text{T}_{3w}\text{S}_s$.

This table demonstrates that our approach, achieved solely by injecting varied levels of randomness into the dataset through controlled data augmentation strength, can attain comparable student performance on both balanced and imbalanced datasets with methods featuring intricate designs on architectures, optimization, or distillation procedures. 

\begin{table}
    \caption{Validation accuracies for our method, LFME, and DMAE on four data sets.}
    \label{tab:val_acc}
    \vskip 0.15in
    \begin{center}
    \begin{small}
    \begin{sc}
    \setlength{\tabcolsep}{1mm}{
    \begin{tabular}{lcccc}
    \toprule
    \multicolumn{1}{c}{Method} & Cifar100 & ImageNet & \begin{tabular}[c]{@{}c@{}}Cifar100\\ imb100\end{tabular} & \begin{tabular}[c]{@{}c@{}}ImageNet\\ long-tail\end{tabular} \\ \midrule
LFME                       & -        & -        & 0.4380                                                        & 0.3880                                                         \\
DMAE                       & \textbf{0.8820}        & \textbf{0.8198}    & 0.3725                                                      & 0.4395                                                        \\
Ours(1T)                   & 0.8133    & -        & 0.5152                                                        & -                                                            \\
Ours(2T)                   & 0.8195    & 0.6917    & 0.5220                                                       & \textbf{0.4968}                                                        \\
Ours(3T)                   & 0.8204        & -        & \textbf{0.5302}                                                        & -                                                            \\ \bottomrule
    \end{tabular}
    }
    \end{sc}
    \end{small}
    \end{center}
    \vskip -0.1in
\end{table}

\section{Model Calibration and Overfitting Effects in our Experiments}\label{app:ECEoverfit}

As a supplementary study, in this section we further investigate the model calibration effects in ensemble KD. Empirically, the student model can be better calibrated by simply enhancing data augmentation strength. And, as the augmentation strength (measured by Affinity) and/or teacher numbers increased, the calibration effects become more pronounced.

While \cite{ICML2017_ECE} has revealed the calibration effects of temperature scaling, a common technique in KD that does not influence the student's accuracy, the impact of data augmentation on the student's prediction confidence and model calibration in KD remains unexplored. This impact is typically gauged by entropy and Expected Calibration Error (ECE) in predictions and is crucial in understanding how they relate to the student's ability to generalize and perform on unseen data, as measured by overfitting tendencies. Our hypothesis is that, beyond the inherent calibration effects of KD, the student model can be effectively calibrated by elevating data augmentation strengths as well.

In this study, we leverage logits entropy and Expected Calibration Error (ECE), along with calibration reliability diagrams \cite{ICML2017_ECE} for visualization, to assess the calibration properties for teachers and student under varied data augmentation strengths. Specifically, the model logits entropy is computed as:

\begin{equation}
    H(\bm{x})=-\sum_{c=1}^{C}\hat{p}(y_c|\bm{x})\log{\hat{p}(y_c|\bm{x})}  
\end{equation}

For ECE calculation, we first group all the validation samples into $M$ interval bins, which are defined based on the prediction confidence of the model for each sample. The ECE thus can be formulated as follows:
\begin{equation}
    \text{ECE}=\sum_{m=1}^{M}\frac{|B_m|}{N} \left | \text{Acc}(B_m) - \text{Conf}(B_m)\right | 
\end{equation}
where $B_m$ denotes the set of samples in the $m$-th bin. The function Acc$(B_m)$ calculates the accuracy within bin $B_m$, while conf$(B_m)$ computes the average predicted confidence of samples in the same bin.

\begin{figure*}[ht]
\vskip 0.2in
\begin{center}
\centerline{\includegraphics[width=\linewidth]{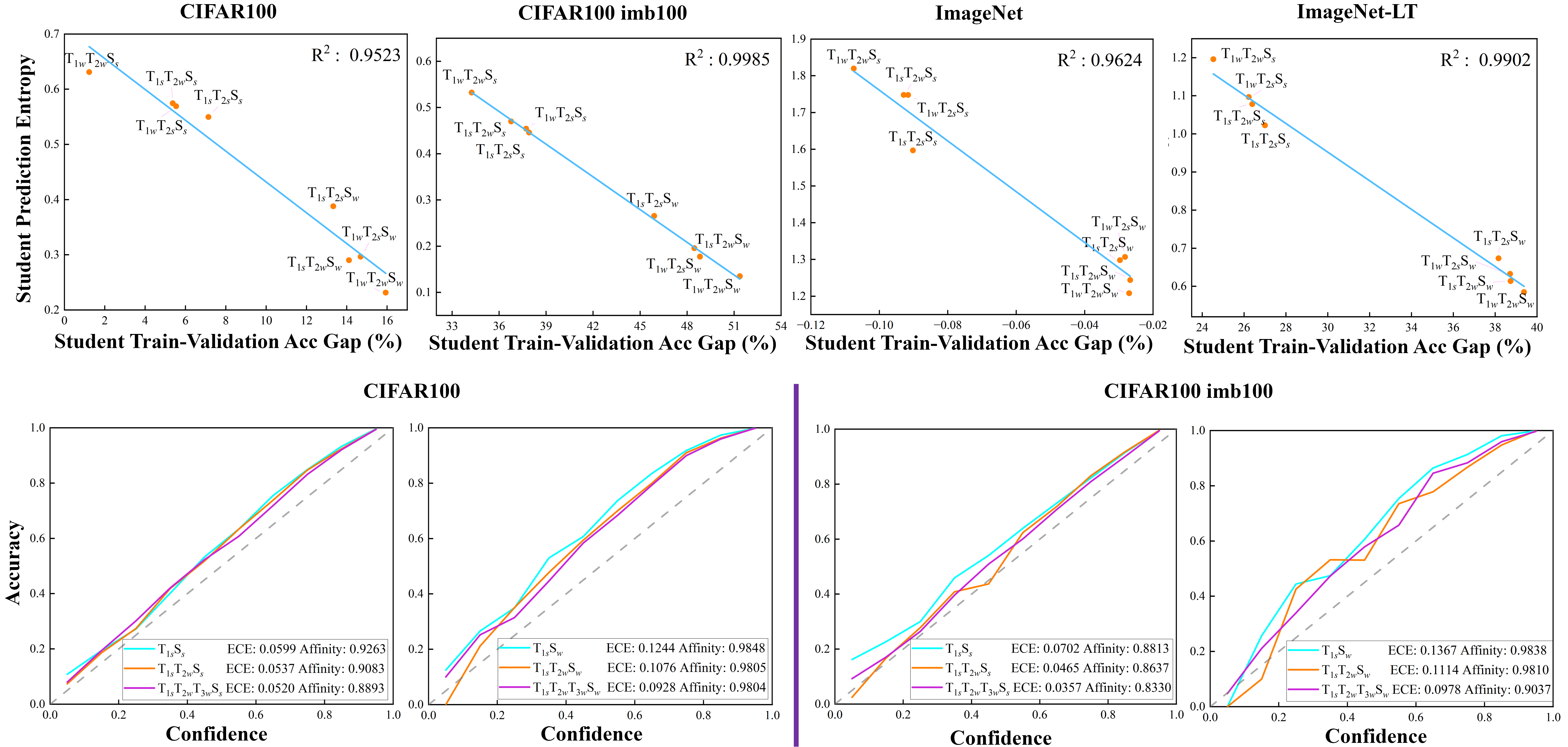}}
\caption{\textit{Top}: Scatter plots of student entropy versus overfitting (gap between top-1 validation and training accuracy) during KD training. \textit{Bottom}: Calibration reliability diagrams with varied teacher numbers (1 to 3) for CIFAR100 imb100 and its balanced counterpart. Stronger augmentation (indicated by decreased Affinity) and more teachers in the ensemble contributes to improved model calibrations and mitigate overfitting effects.}
\label{fig:APP_S_ECE}
\end{center}
\vskip -0.2in
\end{figure*}

In Figure \ref{fig:APP_S_ECE} \textit{Top}, a notable inverse relationship was observed between the entropy of the student model's predictions and overfitting. While stronger data augmentation leading to increased entropy (indicative of lower confidence), there was a concurrent decrease in the tendency of the student model to overfit the training data, as evidenced by the reduction in the train-validation accuracy gap.
Figure \ref{fig:APP_S_ECE} \textit{Bottom} further compares the model calibration reliability diagrams for KD with varied teacher numbers (from 1 to 3) and data augmentation strengths. It can be observed that as the number of teachers increased or the augmentation strength increased (indicated by decreased Affinity), the student models exhibited better calibration.


\begin{table*}[]
    \caption{ECE and Affinity of models with various data augmentation strengths.}
    \label{tab:APP_ECE_Affi}
    \vskip 0.15in
    \begin{center}
    \begin{tiny}
    \begin{sc}
    \setlength{\tabcolsep}{1.05mm}{
    \begin{tabular*}{\linewidth}{lccccccccc} 
    \toprule
    \multirow{2}{*}{Dataset}                                                      & \multirow{2}{*}{Metric} & \multicolumn{8}{c}{Model}                                                             \\ \cmidrule(){3-10}
                                                                              &                         & T$_{1w}$T$_{2w}$S$_w$ & T$_{1w}$T$_{2w}$S$_s$ & T$_{1s}$T$_{2w}$S$_w$ & T$_{1s}$T$_{2w}$S$_s$ & T$_{1w}$T$_{2s}$S$_w$ & T$_{1w}$T$_{2s}$S$_s$ & T$_{1s}$T$_{2s}$S$_w$ & T$_{1s}$T$_{2s}$S$_s$ \\ \midrule
\multirow{2}{*}{Cifar100}                                                     & ECE                     & 0.0776   & \textbf{0.0124}   & 0.1076   & 0.0537   & 0.0994   & 0.0568   & 0.1397   & 0.0745   \\
                                                                              & Affinity                & 0.9807   & \textbf{0.8611}   & 0.9805   & 0.9083   & 0.9858   & 0.9143   & 0.9729   & 0.9310   \\ \midrule
\multirow{2}{*}{\begin{tabular}[c]{@{}c@{}}Cifar100\\ imb100\end{tabular}}    & ECE                     & 0.0979  & 0.0103  & 0.1114  & 0.0465  & 0.0711  & 0.0482  & 0.1303  & 0.0651 \\
                                                                              & Affinity                & 0.9763  & \textbf{0.8132}  & 0.9810  & 0.8637  & 0.9751  & 0.8635  & 0.9723  & 0.8955   \\ \midrule
\multirow{2}{*}{ImageNet}                                                     & ECE                     & 0.0275   & \textbf{0.0095}   & 0.0233   & 0.0118   & 0.0126   & 0.0107   & 0.0122   & 0.0193   \\
                                                                              & Affinity                & 0.9901   & \textbf{0.8767}   & 0.9930   & 0.8988   & 0.9845   & 0.9131   & 0.9871   & 0.9122   \\ \midrule
\multirow{2}{*}{\begin{tabular}[c]{@{}c@{}}ImageNet\\ long-tail\end{tabular}} & ECE                     & 0.0322   & 0.0226   & 0.0357   & 0.0224   & 0.0494   & 0.0307   & 0.0499   & \textbf{0.0178}   \\
                                                                              & Affinity                & 0.9850   & \textbf{0.8311}   & 0.9755   & 0.8704   & 0.9782   & 0.8751   & 0.9903   & 0.8971   \\ \bottomrule 
    \end{tabular*}
    }
    \end{sc}
    \end{tiny}
    \end{center}
    \vskip -0.1in
\end{table*}

Table \ref{tab:APP_ECE_Affi} further provides the Expected Calibration Error (ECE) with corresponding Affinity values for all the trials with 2-teacher ensemble KD. 
This aids in understanding the data augmentation strengths and the decreasing tendencies in all the previous scatter plots: Recall that Affinity measures the offset in data distribution between the original one and the one after data augmentation captured by the student, and lower Affinity corresponds to higher augmentation strength, leading to higher student accuracy. Thus, for the trials with strong data augmentation (e.g., T$_{1w}$T$_{2w}$S$_s$ in CIFAR-100, CIFAR-100 imb100, and ImageNet; T$_{1s}$T$_{2w}$S$_s$ in ImageNet-LT), they not only correspond to a relatively small ECE but also a high validation accuracy.




\bibliographystyle{elsarticle-harv} 


\end{document}